\documentclass[11pt, a4paper]{article}
\usepackage[utf8]{inputenc}
\usepackage[textwidth=15cm]{geometry}
\usepackage{authblk}
\usepackage{pdfpages}
\usepackage{outlines}
\usepackage[autostyle]{csquotes}
\usepackage{array}
\usepackage{caption}
\usepackage{rotating}
\usepackage{color,soul}
\usepackage{float}
\setstcolor{red}
\usepackage{longtable}

\usepackage{placeins}
 \usepackage{apacite}
 \usepackage{hyperref}  % Hyperlinks bib references.
 \usepackage{natbib}
 
\hypersetup{
    colorlinks=true,
    linkcolor=blue,
    filecolor=magenta,      
    urlcolor=cyan,
    citecolor=blue,
}
\usepackage{amsmath}
\usepackage{amssymb}
\usepackage{enumitem}
\usepackage{booktabs}
\usepackage{amsmath}
\usepackage{multirow}
\usepackage[toc,page]{appendix}

 \hypersetup{
    colorlinks=true,
    linkcolor=blue,
    filecolor=magenta,      
    urlcolor=cyan,
    citecolor=blue,
}

\title{Predictive Representativity: Uncovering Racial Bias in AI-based Skin Cancer Detection}
\date{\today}
\author[1I]{Andrés Morales-Forero}
\author[2II]{Lili J. Rueda}
\author[3III]{Ronald Herrera}
\author[1IV]{Samuel Bassetto}
\author[4V]{Eric Coatanea}
\affil[1]{Mathematics and industrial engineering (MAGI), Montreal University (Polytechnic), Montreal, Canada}
\affil[2]{Infectious and Clinical Dermatology Research Group, Universidad El Bosque, Bogotá, Colombia}
\affil[3]{Boehringer Ingelheim International GmbH, Ingelheim am Rhein, Rheinland-Pfalz, Germany}
\affil[4]{Faculty of Engineering and Natural Sciences (ENS), Tampere University, Finland}
\affil[I]{\tt{andres.morales-forero@polymtl.ca}}
\affil[II]{\tt{lijrueda@unbosque.edu.co}}
\affil[III]{\tt{ronald.ferney\_herrera.clavijo@boehringer-ingelheim.com}}
\affil[IV]{\tt{samuel-jean.bassetto@polymtl.ca}}
\affil[V]{\tt{eric.coatanea@tuni.fi}}

\date{} 

\providecommand{\keywords}[1]{\noindent\textbf{\textit{\small Keywords---}} \small #1}

\emergencystretch=\maxdimen
\begin{document}

\maketitle

\begin{abstract}
    Artificial intelligence (AI) systems increasingly inform medical decision-making, yet concerns about algorithmic bias and inequitable outcomes persist, particularly for historically marginalized populations. This paper introduces the concept of \emph{Predictive Representativity (PR)}, a framework of fairness auditing that shifts the focus from the composition of the data set to outcomes-level equity. Through a case study in dermatology, we evaluated AI-based skin cancer classifiers trained on the widely used HAM10000 dataset and on an independent clinical dataset (BOSQUE Test set) from Colombia. Our analysis reveals substantial performance disparities by skin phototype, with classifiers consistently underperforming for individuals with darker skin —despite proportional sampling in the source data. We argue that representativity must be understood not as a static feature of datasets but as a dynamic, context-sensitive property of model predictions. PR operationalizes this shift by quantifying how reliably models generalize fairness across subpopulations and deployment contexts. We further propose an \emph{External Transportability Criterion} that formalizes the thresholds for fairness generalization. Our findings highlight the ethical imperative for post-hoc fairness auditing, transparency in dataset documentation, and inclusive model validation pipelines. This work offers a scalable tool for diagnosing structural inequities in AI systems, contributing to discussions on equity, interpretability, and data justice and fostering a critical re-evaluation of fairness in data-driven healthcare.
\end{abstract}

\section*{Introduction}

Consider a clinical scenario in which a dermatologist employs an artificial intelligence (AI) diagnostic tool, trained primarily on lighter-skinned populations, to evaluate a suspicious skin lesion on a patient with darker skin. The tool classifies the lesion as benign, potentially delaying vital treatment. Although hypothetical, this situation exemplifies a significant concern: biases in training data can systematically disadvantage specific demographic groups, exacerbating rather than mitigating healthcare inequalities.\\

\noindent This issue is particularly alarming given the paradox in skin cancer epidemiology: while melanoma incidence is substantially higher among lighter-skinned individuals— 22 vs 0.9 per 100,000—, darker-skinned patients experience disproportionately higher mortality rates —3.75 times more \citep{gohara2008skin,wu2011racial}— due to delayed or missed diagnoses \citep{culp2019peer,gohara2008skin,wu2011racial}. Factors contributing to these disparities include limited healthcare access \citep{brady2021racial,cortez2021impact}, misconceptions about immunity to skin cancer among darker-skinned individuals \citep{gupta2016skin}, insufficient patient and physician education \citep{kim2009perception,rizvi2022bias}, and persistent racial biases embedded within medical training and literature \citep{epstein2020color,louie2018representations,lester2020absence}. Against this backdrop, a pressing question arises: does AI replicate—and potentially amplify—human biases in dermatology? In this study, we begin to address this question by examining benchmark algorithms trained on a widely used dermatological image dataset: HAM10000 \citep{tschandl2018ham10000}. This dataset has become a cornerstone in the development and benchmarking of machine learning models for skin lesion classification, including in high-profile competitions such as the International Skin Imaging Collaboration (ISIC) challenges. Nevertheless, concerns have emerged regarding the dataset’s lack of demographic diversity, particularly its overrepresentation of lighter skin phototypes and underrepresentation of darker skin tones \citep{morales2024insight}. Such imbalances raise important questions about the fairness and generalizability of models trained on these data, especially when deployed in clinical settings with more diverse patient populations.  \\

\noindent Scrutinizing the predictive performance of machine learning algorithms trained on HAM10000 is challenging because the dataset lacks phenotypic descriptors such as skin phototype. This omission hinders internal equity validation within the source population, particularly with regard to ethnicity or skin tone. Furthermore, in individuals with darker skin, skin lesions frequently appear on lighter areas of the body—such as the palms of the hands or soles of the feet—making it especially difficult to infer a person’s predominant skin tone from dermoscopic images alone. This complicates efforts to assess model fairness across skin tone subgroups and underscores the need for more comprehensive metadata in dermatological datasets. In order to address these issues, we collect a completely independent dataset called BOSQUE Test set \citep{DVN/AQEPIN_2025}: 167 dermoscopic images from Bogotá, Colombia, annotated with lesion types and Fitzpatrick phototypes. We train and evaluate five benchmark CNN architectures —ResNet-50, DenseNet-121, MobileNet-V2, EfficientNet-V2-B0, and VGG-16— on that data set, then compare their demographic parity in the BOSQUE Test set. The results reveal notable disparities in model performance: all five models exhibited reduced precision and recall rates to detect malignant lesions in individuals with darker skin tones (Fitzpatrick types IV–VI) compared to those with lighter skin tones (types II–III). Specifically, the models are less precise when identifying malignant lesions in darker-skinned individuals, meaning they are more likely to misclassify cancerous lesions as benign in this group. This is especially dangerous in clinical settings, as it could delay diagnosis and treatment for those already at higher risk of being underserved.  These discrepancies persisted across architectures, suggesting that the bias is rooted in the training dataset rather than the model design alone. After all, how can a model learn diagnostic nuances specific to darker skin tones if it has never seen enough relevant examples? This issue goes beyond simple demographic representation; it speaks to the necessity of capturing the full spectrum of dermatological presentations across diverse populations to ensure equitable model performance. \\

\noindent Our findings provide evidence that AI systems can replicate biases already embedded in human dermatological practices. More concerning, however, is the potential for these systems to amplify such disparities—particularly when the concept of representativity is used uncritically as a proxy for model performance. For instance, HAM10000 documentation asserts diversity and representativity, citing the inclusion of multiple lesion types and a large sample size. However, HAM10000 disproportionately features lighter skin phototypes, reported to be in a 20:1 ratio compared to darker skin tones —around 5\% \citep{morales2024insight}. Although this might reflect the distribution of the source population —literature reports 22:1 ratio \citep{gohara2008skin,wu2011racial}—, our study shows that representativity based solely on input population proportions is inadequate for contexts involving conditional predictions, such as machine learning. In other words, dataset may accurately reflect the proportions of different subgroups (e.g., demographics, disease types) in the real-world population they were drawn from. However, this does not guarantee good performance when the dataset is used for ML models making conditional predictions (predictions that depend on specific input features, like diagnosing a disease from an image). Therefore, claims of dataset representativity or diversity should be treated with caution, as they may offer a misleading impression of quality with respect to predictive performance.   \\

\noindent Traditional notions of representativity focus primarily on input distributions, stating that statistical findings derived from representative samples should generalize internally to their source population. This assurance relates to \emph{internal transportability} and applies only to specific population parameters and subgroups that have been explicitly validated, and not to predictive parameters. However, AI systems are often deployed beyond their original development contexts, encountering target populations that differ demographically and clinically from the source. As a result, the central challenge shifts from input representativeness to  \emph{external transportability}:  ensuring that AI model predictions remain accurate and equitable across diverse subpopulations in novel, real-world settings.\\

\noindent To address this critical gap, we propose the concept of \emph{Predictive Representativity (PR)}, shifting the focus from input datasets to predictive outputs. Rather than evaluating datasets solely based on how well they mirror population-level input distributions, PR emphasizes the need for model outputs to be accurate and fair across all relevant subgroups. This requires measuring how prediction performance—such as sensitivity, and specificity —varies across demographic strata, particularly those historically underrepresented or disadvantaged in the training data. PR accounts for the conditional nature of machine learning tasks, recognizing that prediction quality is not merely a function of data quantity, but also of data quality and distributional alignment with the target use context.\\

\noindent Moreover, PR explicitly acknowledges the imperfect, context-sensitive nature of real-world ``ground truths'', emphasizing the relative consistency and fairness of predictive performance across demographic groups within both source and intended populations. The PR framework thereby integrates internal validation (equity within source populations) with external validation (equity across new target populations), making fairness evaluation an essential aspect of transportability. \\

\noindent In dermatology, this translates directly into our empirical assessment using HAM10000 and BOSQUE Test set. HAM10000 —sourced from Australia and Austria— lacks explicit metadata on ethnicity or skin tone, precluding internal validation of equitable performance across different skin types. Conversely, BOSQUE Test set—an independently collected, non-probabilistic sample from Bogotá, Colombia—features a higher proportion of darker skin phototypes. Despite its limitations in generalizability to the entire Colombian population, BOSQUE Test set serves as a critical external benchmark — or counter example, uncovering significant racial biases in predictive accuracy. Indeed, our evaluation demonstrates pronounced disparities in model performance, underscoring the inadequacy of proportional dataset composition alone. Moreover, These findings underscore a critical gap in current fairness assessments—while conventional dataset balancing strategies attempt to mitigate bias
during model training, they do not necessarily guarantee fair generalization in
deployment settings.\\

\noindent Ultimately, our analysis highlights the urgent scientific and ethical imperative for fairness-aware dataset curation, targeted data augmentation strategies, and rigorous external validation procedures. Through PR, we advocate for an output-centered audit that ensures AI-driven diagnostic tools deliver equitable healthcare benefits, actively mitigating rather than perpetuating systemic biases and health disparities.\\

\noindent Beyond dermatology, PR has broad implications for AI fairness, interpretability, and regulatory oversight. By reframing representativity as a property of model outputs rather than input data, PR provides a quantifiable and interpretable metric for auditing AI systems across various domains. This perspective not only aligns with ongoing discussions in fairness-aware machine learning but also integrates foundational principles from epidemiology and statistical inference.  PR represents a probabilistic framework for quantifying conditional generalization performance across subpopulations, offering a principled measure of how well a model's predictive behavior aligns with true outcome distributions under covariate and concept shifts. It accounts for both epistemic and aleatoric uncertainty, grounding fairness evaluation in distributional comparisons that reflect real-world deployment contexts rather than idealized training conditions. \\

\noindent  Our study also introduces an \emph{External Transportability Criterion} based on PR, which evaluates whether a model maintains fair and accurate performance when applied to new, demographically different populations. Practically, the External Transportability Criterion complements existing fairness metrics by providing a deployment-focused audit tool that is robust to hidden stratification and demographic drift. It supports pre-deployment risk analysis by flagging subgroup-specific performance gaps that exceed clinically acceptable margins, even when the overall model performance appears satisfactory. In doing so, it operationalizes a probabilistic, task-aware definition of fairness that integrates statistical divergence, clinical relevance, and ethical oversight into a unified metric for real-world AI safety. Moreover, the criterion reflects a key insight from statistical learning theory and causal inference: generalization guarantees must account not only for marginal input distributions but also for conditional label dynamics and structural biases. This includes differences in presentation patterns, labeling practices, and even clinical manifestation of disease across populations—factors that are especially relevant in dermatology and other visually-dependent diagnostic domains.
%\noindent To illustrate the pertinence and necessity of this framework, we present a case study in AI-driven dermatology, where racial disparities in skin cancer detection models highlight the consequences of a misleading claim of representativity. By analyzing the predictive performance of widely used deep learning models on diverse skin tones, we reveal critical gaps that could lead to misdiagnoses and unequal healthcare outcomes. Through the PR concept, we offer a roadmap for improving fairness and reliability in medical AI—ensuring that technological advancements serve all populations equitably.\\

%At a time when AI is reshaping society at an unprecedented pace, the need for safe models has never been more urgent. This paper lays the foundation for a new standard in algorithmic fairness—one that goes beyond dataset composition to interrogate the very structure of predictive systems. By doing so, we take a crucial step toward more ethical, accountable, and effective AI.
\section*{Background}

During the past century, the concept of representativity has evolved from an essential methodological principle in statistical inference to a multifaceted concept that touches disciplines as diverse as cognitive psychology, political science, and, most recently, machine learning. Early twentieth-century statistical treatises framed representativity primarily in terms of sample selection: ensuring that a studied subgroup mirrored the broader population with sufficient fidelity to draw valid, generalizable conclusions \citep{kish1965survey}. These foundations—probabilistic sampling techniques, consideration of demographic strata, and variance minimization—formed the bedrock of empirical research in the social and natural sciences.\\

\noindent By the 1970s, the concept was further nuanced by cognitive psychology. \cite{tversky1974judgment} introduced the representativeness heuristic to describe how individuals judge likelihood based on similarity to a prototype rather than strict statistical rules. This shift underscored that representativity is not only a property of data or samples but also a cognitive shortcut, prone to bias. It broadened the conversation from purely methodological protocols — “How do I create a representative sample?”— to the psychological processes that shape our perceptions of what is, in fact, “typical”. \\

\noindent This psychological perspective raises philosophical questions about how decisions and representations are shaped. In existentialist and humanistic traditions (e.g., \cite{merleau1962phenomenology}, \cite{maslow1943theory}, \cite{rogers1959theory}), perception is seen as an active, context-driven process. Representativity, then, is not an absolute metric but relative to the observer’s intent and context. In machine learning, models reflect and reinforce specific values rather than merely capturing data. Fairness-aware systems must therefore acknowledge the biases in both data and ground truth, aiming not to erase subjectivity but to engage with it ethically and transparently.\\

\noindent Epidemiology has long debated the necessity and validity of representativity in scientific research. While some argue that representative sampling is crucial for external validity \citep{elwood2013commentary}, others caution against its overuse, emphasizing that internal validity should take precedence \citep{richiardi2013commentary}. \cite{rothman2013rebuttal} contend that representativeness should be avoided when conducting causal inference studies, as scientific generalization does not rely on sampling representativity but rather on controlling confounding variables and understanding causal mechanisms. They argue that while representative samples may be necessary for descriptive epidemiology and public health research, they can often be unnecessary or even misleading in causal studies \citep{rothman2013representativeness}.\\

\noindent Additionally, scholars such as \cite{ebrahim2013commentary} have pointed out that non-representative cohorts may lead to biased exposure-outcome associations. They highlight that selection bias and confounding factors can significantly distort causal inferences if representativity is not properly considered. Similarly, \cite{swanson2012uk} discusses selection bias in large-scale epidemiological studies like the UK Biobank, arguing that low participation rates (e.g., 5.5\% in UK Biobank) may challenge the generalizability of findings. However, \cite{rothman2013rebuttal} counter that the validity of epidemiological research is better ensured by careful study design rather than insistence on representativity. \\

\noindent In recent decades, the digital revolution and the ascendancy of ML have dramatically expanded the reach —and stakes— of representativity. With algorithms now underlying decisions about finance, healthcare, and public safety, ensuring that data and models adequately capture diverse population attributes has become crucial \citep{barocas2023fairness}. Inadequate representativity in training data can lead to biased models with real-world consequences, including discrimination and systematic exclusion. Beyond traditional sampling concerns, ML researchers now grapple with complexities of dynamic data streams, “long-tail” phenomena, and multidimensional fairness criteria, all of which demand new frameworks for maintaining—and measuring—representativity \citep{bender2018data}.\\

\noindent Building upon these expanded demands, a new wave of frameworks and methods has emerged to ensure that modern data-driven systems address representativity in both their inputs and outputs. Researchers have proposed data-centric strategies— such as augmenting underrepresented classes \citep{chawla2002smote}, generating synthetic data through generative adversarial networks \citep{goodfellow2014generative}, and systematically documenting dataset composition via “datasheets” or “model cards” \citep{gebru2021datasheets,mitchell2019model}— to identify and mitigate coverage gaps. Additional approaches, including active sampling and transfer learning, enable models to continually recalibrate their understanding as they encounter new data or shift into related domains.\\

\noindent On the algorithmic side, a variety of fairness-aware techniques have been developed to detect and correct biased outcomes. Methods range from pre-processing approaches (rebalancing data distributions) to in-processing mechanisms (e.g., adversarial debiasing) and post-processing adjustments \citep{hardt2016equality,feldman2015certifying}. These frameworks account for multidimensional fairness criteria —including notions of demographic parity, equalized odds, and counterfactual fairness— to better capture the complexity of real world group boundaries and intersectional identities.
Recent work has also explored trade-offs between fairness and accuracy, highlighting that different fairness metrics (e.g., disparate impact, equalized odds, calibration, and statistical parity) may lead to conflicting outcomes in model evaluation and deployment \citep{corbett2017algorithmic, mehrabi2021survey}.
In contexts where data distributions change over time, online learning and domain adaptation provide ways to continually update model parameters, ensuring that once -representative training sets do not ossify into biased solutions.\\

\noindent Despite the long history and widespread use of the concept of representativeness, there is still no consensus on its formal definition. In classical statistics, some frame representativity as a condition in which each member of a population has a nonzero probability of being included in a sample \citep{kish1965survey}. \cite{tille2006sampling} takes a more stringent position, suggesting that a sampling strategy (the pairing of the probability distribution of all possible samples and the estimator) is deemed representative if it  unbiased estimation of a total with zero variance. With the advent of big data and ML, the notion of representativeness has grown even more expansive.  \cite{kruskal1979I,kruskal1979II,kruskal1979III,kruskal1980representativeIV} provided a seminal framework by identifying six key notions that underlie the idea of a “representative sample”: assertive acclaim, emphasizing unwarranted claims of representativity; absence of selective forces, which highlights unbiased sampling; miniature population, representing proportional distributions of subpopulations; typical observation, capturing the average or ideal case; coverage, ensuring all subpopulations are included; and methodological rigor, describing adherence to systematic sampling methods. Subsequently, \cite{clemmensen2022data} expands on these notions within contemporary ML contexts adding two elementes in scientific AI writing: “copycat,” which underscores the generation of synthetic data that accurately reflect target population characteristics, and “no notion,” which indicates the absence of clear criteria for representativity or acknowledgment of data limitations.\\

\noindent Among Kruskal’s six notions, coverage is particularly significant for fairness applications because it emphasizes inclusion of all subpopulations, even those with a lower natural prevalence. This can involve intentionally oversampling or otherwise augmenting underrepresented groups to ensure their representation is robust enough to inform model training and evaluation. Although this approach may deviate from purely natural distributions, and thus potentially reduce “statistical” alignment, it aims to improve generalizability of results in different demographic segments. However, as \cite{clemmensen2022data} note, a substantial percentage of NeurIPS (61.1\%) and ICCV (84.4\%) publications adopt some form of coverage strategy without fully addressing the attendant resource demands or the trade-offs in model bias and variance.\\

\noindent Some years earlier, \cite{grafstrom2014select} had introduced a definition of representative samples as a scaled-down and well-spread version of the population that robustly captures essential characteristics. Through simulations, this approach demonstrated notable reductions in variance and improved spatial balance. Nonetheless, its reliance on known inclusion probabilities constrains its applicability to probability-based samples— an assumption often at odds with observational data commonly used in ML. Consequently, adapting such definitions to nonprobabilistic or large-scale machine-learning scenarios remains a pressing challenge, underscoring the continued evolution of representativity as an integrative, cross-cutting concept. \\

\noindent Additional recent work has re-examined the boundaries and implications of representativeness across multiple domains. For example, \cite{rudolph2023defining} defines a sample as representative if findings derived from it can be generalized to the intended target population—an idea deeply relevant to medical and public health research where external validity is paramount. This resonates with \cite{lavrakas2008encyclopedia} meaning on Representative Samples in the Encyclopedia of Survey Research Methods, which equates representativeness with external validity: a truly representative sample must reflect the population so that conclusions can legitimately extend beyond the sampled individuals.\\

\noindent The connection between a sample and its external validity highlights the purpose-driven nature of representativity, emphasizing that ``representativeness'' depends on whether a sample accurately captures the broader population relevant to the specific research question or task at hand. 
\cite{rothman2013representativeness} and \cite{richiardi2013commentary} argue that scientific inference does not inherently require representativity. Instead, inference depends on methodological rigor, proper study design, and statistical adjustments rather than simple notions of representativeness. This shift reorients the discussion from ``representative samples'' to a more comprehensive understanding of ``representativity'', which extends beyond the samples themselves to encompass indentifying meaningfull examples, emploing effective estimators and utilizing robust predictive models. This evolving definition broadens the scope of representativity toward integrating both data acquisition and model performance, ensuring that both the sample and the methods used to analyze it are capable of making valid, generalizable inferences. \\

\noindent  The dynamic nature of ML tasks necessitates approaches distinct from cross-sectional studies, shifting towards a more scientific perspective. While cross-sectional studies rely on a descriptive analysis of the population, requiring sufficient samples and unbiased, low-variance estimators, ML has its own demands. ML requires conditional predictions and hypothesis testing over population-level descriptions. This shift moves the emphasis away from traditional notions of representativeness toward a more broader concept of representativeness. ML must address distinct research questions that demand methods and data beyond those used solely for training. This includes considerations of data dynamics, temporal dependencies, and evolving distributions that are often overlooked in static analyses. Unlike traditional studies that assume fixed population characteristics, ML must also account for changing conditions, social context, feedback loops, and real-time decision-making processes.

%\noindent  Furthermore, ML research questions often involve causal inference, generalization across domains, and robustness under varying conditions, requiring novel experimental designs and data collection strategies. This necessitates a blend of observational and experimental data, adaptive sampling techniques, and an iterative approach to model refinement.

%Meanwhile, What exactly is a 'representative sample'? (2023) underscores the purpose-driven nature of representativeness, emphasizing that “representativeness” hinges on whether a sample usefully captures the population for the question or task at hand. In more theoretical or critical discussions, Representative Sampling, I: Non-Scientific Literature observes how labeling a sample “representative” often conveys trust—implicitly assuring the reader that the sample will not mislead them, though the basis for this assurance can be uncertain or context-dependent.

%Turning back to machine learning and AI, Data Representativity for Machine Learning and AI Systems (2022) conducts a broad survey of representativity definitions, highlighting measurable constructs—such as coverage and proportional representation—that can help practitioners evaluate different data samples. Lastly, What does it mean to be ‘representative’? (2022) proposes a comprehensive taxonomy of how representativity can be interpreted in diverse empirical settings, effectively bridging the gap between population-based and use-case-specific views.

\section*{Theoretical Framework of Predictive Representativity}

\noindent Across disciplines, the term \textit{representative sample} is used in at least three complementary senses highlighted in the literature review:

\begin{enumerate}[label=(\roman*)]
    \item the \emph{probabilistic} view, in which every unit has a known, non-zero inclusion probability so that sample frequencies mirror population frequencies \citep{kish1965survey};  
    \item the \emph{coverage/miniature-population} view, which stresses that all relevant subgroups must be sufficiently captured, even if this requires deliberate oversampling \citep{kruskal1979I,kruskal1979II,kruskal1979III,kruskal1980representativeIV}; and
    \item the \emph{purpose-driven} view from epidemiology and causal inference, where a sample is “representative” only insofar as it supports the specific descriptive or causal question at hand \citep{tille2006sampling,rothman2013representativeness}.  
\end{enumerate}

\noindent When any of these criteria are satisfied, internal transportability is supposed to be assumed: estimates computed on the sample generalize to the source population.\\

\noindent Machine-learning deployment poses a harder question: Will a model trained on the population $\Pi$ behave equitably and reliably when applied to a different population $\Pi'$? Covariate shifts, label noise, and systemic bias imply that classical representativity—even if perfectly realized in $\Pi$—says nothing about performance in $\Pi'$.\\

\subsection*{Modelling set-up}

In the context of supervised learning, we begin by defining the fundamental random variables involved in the process. Let \( X \in \mathcal{X} \) represent the input features, and \( Y \in \mathcal{Y} \) denote the corresponding true label. These variables are drawn from some underlying joint distribution, capturing the relationship between the inputs and their associated outputs.\\

\noindent A predictive model or estimator is defined as a function \( A : \mathcal{X} \to \mathcal{Y} \), which maps input features to predicted labels. The prediction produced by this estimator is denoted by \( \hat{Y} = f_A(X) \), where \( f_A \) encapsulates the decision rule learned by the model based on training data.\\

\noindent To formalize generalization across different contexts, we consider two populations: the source population \( P \) and the target population \( P' \). These are characterized by the joint probability distributions over \( (X, Y) \) in their respective domains. Understanding how the model performs across these distributions is critical for evaluating its robustness and transferability.\\

\noindent Furthermore, we can analyze behavior within specific regions of the input space by defining sub-populations. For any measurable subset \( S \subseteq \mathcal{X} \), the sub-population distribution \( P_S \) refers to the conditional law of \( (X, Y) \) given that \( X \in S \). The conditional distribution of labels given an input \( x \) under this sub-population is written as \( P_S(Y \mid X = x) \). Correspondingly, the model’s prediction in this region is represented by the degenerate distribution \( \hat{P}_S(Y \mid X = x) = \delta_{f_A(x)} \), where \( \delta_{f_A(x)} \) denotes a point mass centered at the model’s output.

%\subsection*{Divergence as a Fairness Perspective}

%To quantify fairness, we evaluate the misalignment between the true label distribution and the model’s predicted distribution using statistical divergences. These divergences, denoted by \( D(\cdot \| \cdot) \), can be chosen from a family of standard measures such as Kullback–Leibler, Jensen–Shannon, total variation, among others. A key property of these divergences is that \( D(p \| q) = 0 \) if and only if the two distributions \( p \) and \( q \) are identical. This framework enables a principled comparison of predictive behaviors across different groups or conditions.

\subsection*{Predictive Representativity}

\textit{Predictive Representativity} is a diagnostic metric designed to quantify the relative misalignment between a model’s predictions and the true outcome distribution within a specific sub-population, in contrast to its behavior over the entire population. It provides a means to assess whether a model generalizes its predictive performance equitably across distinct demographic or contextual subgroups.\\

\textbf{Definition}\\
Let \( A \) be a fixed predictive model, \( P \) the source population, and \( S \subseteq \mathcal{X} \) a measurable sub-population of interest. \emph{Predictive Representativity (PR)} is formally defined as:

\[
\mathrm{PR}(P, S, A) = \mathbb{E}_{X \sim P_S} \left[ D \left( P_S(Y \mid X) \,\|\, \hat{P}_S(Y \mid X) \right) \right] 
- \mathbb{E}_{X \sim P} \left[ D \left( P(Y \mid X) \,\|\, \hat{P}(Y \mid X) \right) \right],
\]

\noindent
where \( D(\cdot \| \cdot) \) is a statistical divergence (e.g., Kullback–Leibler, Jensen–Shannon, or total variation), and \( \hat{P} \) represents the predictive distribution output by model \( A \). A positive PR value indicates worse performance in the sub-population \( S \) relative to the population average, while a negative value may suggest overfitting or unfair optimization for \( S \) (see Table \ref{tab:pr_interpretation}). The first term captures the divergence between the ground truth and predictions within the sub-population \( S \), while the second reflects this divergence over the full population.\\

\begin{table}[ht]
\centering
\footnotesize
\caption{Interpretation of Predictive Representativity (PR) values.}
\begin{tabular}{@{}cl@{}}
\toprule
\textbf{Value of PR} & \textbf{Interpretation} \\ 
\midrule
\multirow{2}{*}{$= 0$} & Predictive divergence in \( S \) matches the model’s average \\
& performance over the entire population. \\
$> 0$ & The model underperforms in \( S \) (predictive inequity, potential harm). \\
$< 0$ & The model overfits or is disproportionately optimized for \( S \). \\
\bottomrule
\end{tabular}
\label{tab:pr_interpretation}
\end{table}

\noindent
In typical model development workflows, global divergences—such as those tied to average prediction error—are minimized through practices like cross-validation, early stopping, and hyperparameter tuning on held-out test sets. As a result, the second term in the PR expression (i.e., divergence over the full population \( P \)) is often very close to zero, bounded near the statistical noise floor. However, when a model is deployed in a new target population \( P' \), this assumption may no longer hold. Distributional shifts, latent confounding factors, or demographic disparities can induce significant deviations in both global and local divergences. Therefore, in deployment scenarios, both components of PR must be re-evaluated, as we demonstrate in our case study using an entirely independent dataset.

\subsection*{External Transportability Criterion}

A model trained on a source population \( \mathcal{P} \) is said to be \emph{transportable} to a different domain \( \mathcal{P}' \) with respect to a sub-group \( S' \) if the PR difference remains within a predefined stakeholder tolerance \( \varepsilon \). Formally, assuming the model \( A \) performs well overall in \( P' \) , transportability is satisfied when
\[
\left| \mathrm{PR}(\mathcal{P}', S', A) \right| \leq \varepsilon.
\]
This condition provides a formal way to assess whether the fairness and accuracy characteristics of a model generalize acceptably across domains. Practically, transportability poses the question: \emph{``Does the fairness/accuracy gap we saw at home remain acceptably small abroad?''}\\

\begin{figure}[H]
\centering
 \includegraphics[width=\textwidth]{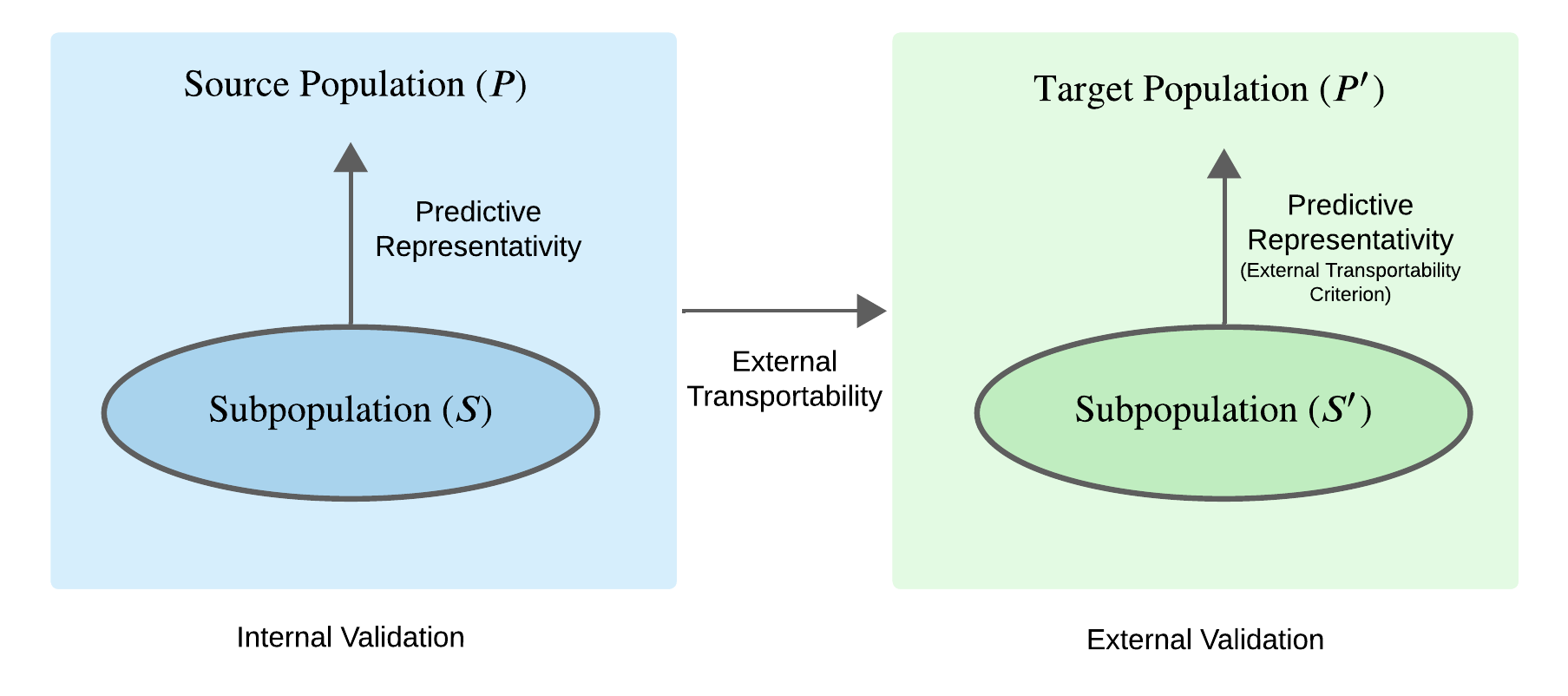}    \\
  \caption{Predictive Representativity and External Transportability Criterion. The model is trained and internally validated in the source domain $\mathcal{P}$, with respect to subpopulation $S$, and then evaluated for fairness transfer to the target domain $\mathcal{P}'$ and subpopulation $S'$. External transportability holds if the performance gap does not exceed a stakeholder-defined tolerance $\varepsilon$.}
\label{fig:PR_diagram}
\end{figure}

\noindent Figure~\ref{fig:PR_diagram} illustrates this concept. On the left, predictive representativity is calculated within the source population $\mathcal{P}$ for a subpopulation $S$, establishing a baseline measure of fairness through internal validation. The model is then deployed in a target domain $\mathcal{P}'$, where the same subpopulation—now represented as $S'$—is re-evaluated to determine whether fairness properties generalize. External transportability is achieved if the predictive representativity in the target population remains within a tolerable deviation $\varepsilon$ from the source. This visual representation underscores that fairness is not guaranteed by training-time validation alone but must be explicitly reassessed when models cross demographic or contextual boundaries.

\subsection*{Empirical Estimation}

To estimate predictive representativity from data, we use a labeled test set \( \{(x_i, y_i)\}_{i=1}^n \subset \mathcal{P} \). The empirical estimate \( \widehat{\mathrm{PR}} \) is computed as the difference in average divergence between the sub-population \( S_I \) and the full population:
\[
\widehat{\mathrm{PR}} = \frac{1}{n_S} \sum_{x_i \in S_I} D\left( \delta_{y_i} \,\|\, \delta_{f_A(x_i)} \right) - \frac{1}{n} \sum_{i=1}^n D\left( \delta_{y_i} \,\|\, \delta_{f_A(x_i)} \right),
\]
where \( \delta_{y_i} \) denotes the point mass at the true label and \( f_A(x_i) \) is the model prediction. Confidence intervals for this estimate can be obtained using non-parametric bootstrap methods.

\subsection*{Connection to Robustness and Fairness}

Predictive Representativity serves as a diagnostic tool in understanding both fairness and robustness of ML models. From a \textbf{fairness} perspective, PR reveals disparities in error rates across protected or sub-population groups, even in cases where the dataset appears demographically balanced.\\ 

\noindent In particular, PR provides an important contrast to a traditional statistical fairness definitions like Demographic Parity (see also the Appendix \ref{appendix:PR_comparisson}). While this metric requires that the output distribution of a model is invariant across groups —formally, \( \hat{Y} \perp S \), where \( S \) denotes a sensitive attribute—, PR evaluates whether the model's performance is consistent across those same groups. \\

\noindent Formally, Demographic Parity is defined as:
\[
\mathbb{P}(\hat{Y} = 1 \mid S = a) = \mathbb{P}(\hat{Y} = 1 \mid S = b), \quad \forall a, b \in \mathcal{S},
\]

\noindent which enforces group-level output parity but does not account for ground truth labels \( Y \). Therefore, to validate whether this parity is \textit{predictively meaningful}, compute PR for each group \( S = a,b\). If:

\[
\mathrm{PR}(\mathcal{P}, S=a,A) \approx  \mathrm{PR}(\mathcal{P}, S=b,A) \approx 0,
\]

\noindent then the model not only outputs predictions equally, but does so with comparable accuracy and alignment to the true label distribution —as we do it in our case study. \\

\noindent Regarding \textbf{robustness}, large absolute values of Predictive Representativity (\( |\mathrm{PR}| \)) indicate a model's failure to generalize consistently across sub-populations. Such inconsistencies may arise due to \textit{covariate shift} (changes in the distribution of input features) or \textit{concept shift} (changes in the relationship between inputs and outputs) between the training and evaluation contexts.\\

\noindent While robustness is a broad property that refers to a model's resilience to various types of perturbations—including adversarial noise, corrupted data, or out-of-distribution inputs—PR captures a specific kind of robustness: \textit{group-conditional prediction stability}. That is, PR provides a framework to evaluate whether the model maintains consistent predictive behavior across socio-demographic or context-defined subgroups.\\

\noindent In this sense, high values of \( |\mathrm{PR}| \) serve as a signal of \textit{latent fragility}: the model may exhibit strong average performance, yet break down when applied to structurally different segments of the input space. Thus, PR acts as a valuable robustness diagnostic, complementing traditional measures of accuracy or calibration by focusing on fairness-aware generalization under real-world distributional variability.\\

\noindent On the other hand, the observations about \( \mathrm{PR} < 0 \) and \( \mathrm{PR} > 0 \) reflect core insights from the \textit{No Free Lunch Theorem} in machine learning. The theorem states that no single model can be optimal across all possible data distributions, implying that some level of trade-off between performance in different subpopulations is inevitable. In the context of Predictive Representativity, a value \( \mathrm{PR}(P, S, A) > 0 \) indicates underperformance in subpopulation \( S \), while \( \mathrm{PR} < 0 \) suggests the model may be disproportionately optimized for that group. These outcomes are not inherently indicative of unfairness, but rather of how performance is distributed across the input space. Improving predictive fairness for one subgroup can—depending on the model, data, and task—decrease performance in others, especially when capacity or data coverage is limited. Such trade-offs must be interpreted in terms of application-specific goals, stakeholder risk tolerances, and ethical priorities. For example, in high-stakes applications like medical diagnostics or lending, prioritizing performance parity in underserved groups may be essential, even at the expense of global accuracy.\\

\noindent Ultimately, these tensions reinforce the need for \textit{context-aware fairness frameworks}, capable of surfacing and quantifying trade-offs rather than concealing them. Predictive Representativity offers a structured way to make these trade-offs transparent. Its outputs enable iterative evaluation strategies where fairness is monitored, adjusted, and refined based on evolving deployment contexts and stakeholder feedback. This emphasizes that fairness in machine learning is not a static property but a dynamic, negotiated process shaped by both technical and societal factors.

\subsection*{Operationalising Predictive Representativity}

Estimating the full conditional distributions \( P(Y \mid X, S) \) and \( \hat{P}(Y \mid X, S) \), which appear in the formal definition of predictive representativity \( \mathrm{PR}(P, S, A) \), is often infeasible in practice. Real-world datasets are finite, subject to noise, and frequently influenced by historical or measurement bias. As a result, the theoretical definition of PR must be approximated using observable proxies that enable practical validation or regulatory checks.\\

\noindent To make PR computable, we introduce a collection of domain-relevant performance metrics and \emph{metric-level predictive representativity} on sub-population \( S \) as follows:\\

\textbf{Definition}\\
\noindent Let \( \mathcal{P} \) denote the population, \( \mathcal{S} \subseteq \mathcal{P} \) denote a subpopulation of interest, and \( \mathcal{A} \) denote a predictive model. Assume the model \( \mathcal{A} \) achieves satisfactory performance on \( \mathcal{P} \), as measured by a set of predefined metrics \( \{\mathcal{M}_1, \mathcal{M}_2, \dots, \mathcal{M}_k\} \), identified by domain experts. We say that a model \( A \) satisfies \emph{metric-level predictive representativity} on sub-population \( S \) if:
\[
|\text{PR}_{\mathcal{M}_i}(\mathcal{S})| \leq \epsilon_i, \quad \forall i \in \{1, 2, \dots, k\},
\]
where:  
\[
\text{PR}_{\mathcal{M}_i}(\mathcal{S}) = \mathcal{M}_i(\mathcal{S}) - \mathcal{M}_i({\mathcal{P}}),
\]  
and \( \epsilon_i \geq 0 \) is a predefined threshold for acceptable deviation, determined by domain experts.\\

\textbf{Example}\\
\noindent \textit{Medical Diagnosis for Disease Detection:} Consider a predictive model \( \mathcal{A} \) developed to identify the presence of a particular disease within a general population \( \mathcal{P} \). Within this population, a sub-group \( \mathcal{S} \) may be defined based on a distinctive genetic predisposition that influences disease susceptibility. To evaluate the model's performance across groups, domain experts focus on two critical metrics: sensitivity, denoted \( \mathcal{M}_1 \), which captures the model’s ability to correctly detect true disease cases; and specificity, denoted \( \mathcal{M}_2 \), which ensures minimization of false positive diagnoses.\\

\noindent To achieve predictive representativity, the model’s sensitivity and specificity on the sub-population \( \mathcal{S} \) should be comparable to its performance on the overall population \( \mathcal{P} \). Formally, this means the differences in these metrics between the two groups must remain within two acceptable tolerance levels \( \epsilon_1 \) and \( \epsilon_2 \), respectively:
\[
\left| \mathrm{PR}_{\mathcal{M}_1}(\mathcal{S}) \right| \leq \epsilon_1 \quad \text{and} \quad \left| \mathrm{PR}_{\mathcal{M}_2}(\mathcal{S}) \right| \leq \epsilon_2.
\]

%The operational definition of predictive representativity provides a critical framework for quantifying model fairness and robustness through performance metrics that are meaningful to practitioners. By evaluating whether key metrics like sensitivity or specificity remain consistent across sub-populations, this formulation allows for practical, data-driven assessments of equity in real-world deployments. This framework becomes especially salient in high-stakes domains such as medical diagnosis, where disparities in model performance can translate into life-threatening consequences.\\

\noindent A striking application of this concept is illustrated in the case study on AI-based skin cancer detection, where classifiers trained on the widely-used HAM10000 dataset show markedly lower predictive representativity for individuals with darker skin tones. By computing metric-level PR values—including precision, AUC-PR, and F1-score—the analysis reveals statistically significant performance gaps. These findings underscore the critical role of PR in uncovering hidden inequities: despite acceptable aggregate accuracy, the classifiers fail to generalize fairly across demographic lines. In this context, PR is not merely a theoretical tool but a vital instrument for guiding ethical model validation, dataset auditing, and the design of more inclusive AI systems.
\section*{Case Study: Uncovering Racial Disparities in AI-Based Imaging Classifiers for Skin Cancer Detection}  

\noindent Skin cancer, particularly melanoma, is among the deadliest cancers globally. While more common in lighter-skinned individuals, those with darker skin often face worse outcomes due to delayed diagnoses \citep{merrill2016worldwide,brady2021racial}. These delays result from limited public awareness, inadequate medical education, and systemic healthcare biases. Advances in AI have produced high-performing imaging classifiers for pigmented skin lesions. However, limited representation of darker skin tones in training data raises concerns about fairness and generalizability \citep{morales2024insight}. Without capturing the complexity of underrepresented groups, such models risk reinforcing health disparities. The HAM10000 dataset—containing 10,015 images across seven lesion types—has been widely adopted in dermatological AI research \citep{tschandl2018ham10000}. Yet, it lacks ethnicity metadata and omits lesion types common in darker skin, restricting its clinical relevance. Although skin cancer is less prevalent in darker-skinned populations, excluding them from model development is unjustified. Underrepresentation can lead to worse diagnostic outcomes, and low prevalence does not imply low clinical importance. Ensuring representativity remains crucial.\\

\begin{figure}[h!]
\centering
 \includegraphics[width=\textwidth]{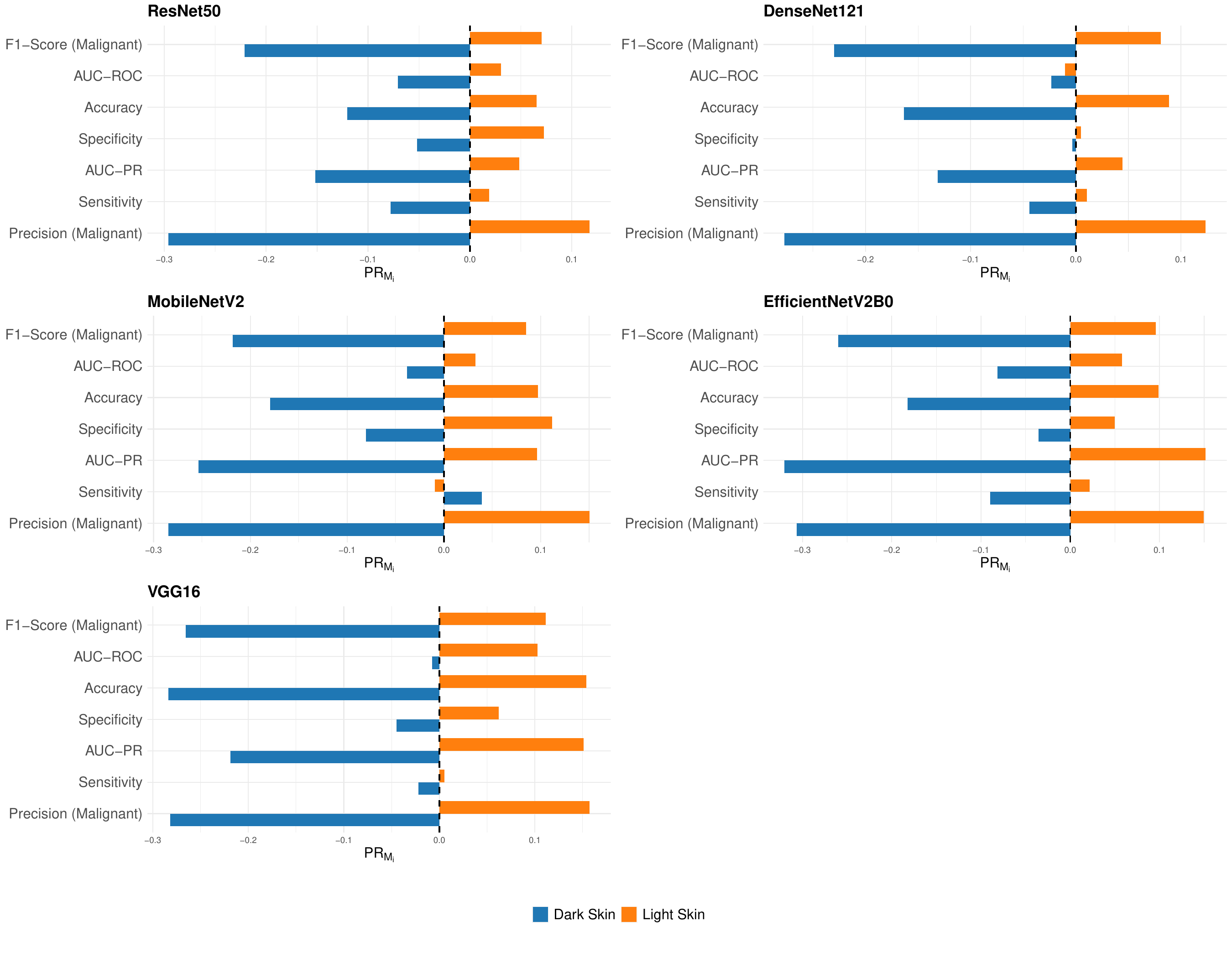}    \\
\caption{Predictive Representativity metrics by model and skin type group.}
\label{fig:PRms}
\end{figure}

\noindent To evaluate HAM10000's representativity for darker skin in malignancy detection, we curated the BOSQUE Test set—165 dermoscopic images from Bogotá, Colombia, annotated with lesion types and Fitzpatrick phototypes. Some diagnoses were biopsy-confirmed. BOSQUE Test set serves as an external benchmark. We trained several deep learning models solely on HAM10000 and applied standard data augmentations (e.g., rotation, scaling, lighting) to reflect clinical variability. Generative methods like artificial skin darkening were excluded due to their inability to reproduce the complex dermatological features of darker skin, including unique textures, pigmentation patterns, and lesion presentations. Instead of improving representativity, such manipulations risk introducing artifacts that distort model training and undermine the validity of performance metrics, ultimately compromising the reliability of the classifiers. The seven classes of HAM10000 lesions were collapsed into binary categories: benign (e.g., nevi, keratoses) and malignant (e.g., melanoma, basal cell carcinoma, actinic keratoses). Oversampling and class weighting were used to address class imbalance. Models were then evaluated on the BOSQUE Test set, with phototypes grouped into lighter (I–III, n=107) and darker (IV–VI, n=58) categories. Using the Predictive Representativity (PR) framework, we assessed model performance across both groups. Our results show clear disparities between lighter and darker skin tones (see Table~\ref{tab:metrics_performance} and Figure~\ref{fig:PRms}). Below, we detail these findings across key performance metrics.

\begin{itemize}
    \item Precision (Malignant):  While the overall accuracy on the new test set aligns closely with the performance observed on the HAM10000 test set (approximately 0.720), precision—a critical metric in this context—reveals the most pronounced disparity in representativity between skin tone groups. Precision PR metrics are consistently lowest for the darker skin group across all models, with the higher precision values from the lighter skin group driving the overall performance upward. For instance, ResNet50 achieves a precision of 0.897 for lighter skin phototypes, compared to just 0.484 for darker skin phototypes. This gap is statistically significant (p $\leq 0.001$), underscoring that while models perform effectively in identifying malignant lesions in lighter skin, their precision drops substantially for darker skin. The positive PR values for the lighter skin group indicate that this subgroup is the best represented for predicting malignant lesions, contributing disproportionately to the overall precision metric. This pattern is consistently observed across all models evaluated. \\
    \item Sensitivity (Recall): The models generally exhibit high sensitivity for detecting malignant lesions across both skin tone groups, though there are subtle differences. The sensitivity for darker skin phototypes tends to be slightly lower than for lighter skin phototypes in most models. For example, ResNet50 shows a sensitivity of 0.789 for darker skin, compared to 0.886 for lighter skin, although this difference is not statistically significant (p = 0.095). This implies that the models maintain good ability to detect malignancies in both skin tones, but slight performance losses for darker skin might occur. This also suggests that sensitivity is less affected by representativity gaps than precision.
    \item AUC-PR (Area Under Precision-Recall Curve): The AUC-PR values capture the models' ability to balance precision and recall for malignant lesion detection. For ResNet50, the PR for lighter skin is 0.048, while for darker skin, it is -0.152, indicating a marked reduction in performance for the latter group. This negative PR for darker skin underscores the model's diminished effectiveness in maintaining a balance between precision and recall, highlighting challenges in achieving equitable predictions across demographic groups. The Z-statistic (3.762, p $\leq 0.001$) confirms the statistical significance of this disparity. Notably, ResNet50 achieves an AUC-PR of 0.749 for darker skin tones, significantly lower than the 0.949 achieved for lighter skin, further emphasizing the model's poorer performance for darker skin in managing both false positives and false negatives.
    \item Specificity: The specificity values, which measure the ability of the models to correctly identify benign lesions, are generally lower for darker skin phototypes. ResNet50 exhibits a PR of 0.072 for lighter skin and -0.052 for darker skin, reflecting a modest gap. However, this difference is not statistically significant (p = 0.104), implying that specificity performance is relatively balanced but still skewed in favor of lighter skin. This suggests that while the models are fairly effective at distinguishing benign lesions in lighter skin phototypes, they struggle more with this task in darker skin phototypes.
    \item Accuracy: Models trained on HAM10000 show a notable drop in accuracy when tested on darker skin tones. The PR values for accuracy further highlight representativity gaps. For instance, ResNet50’s PR is 0.065 for lighter skin and -0.121 for darker skin. The significant negative PR for darker skin (Z = 2.734, p = 0.006) indicates that overall model accuracy disproportionately favors lighter skin tones, aligning with the trends seen in other metrics.\\
    \item F1-Score (Malignant): The F1-score, which balances precision and recall, also reveals a clear performance gap. ResNet50 achieves an F1-score of 0.892 for lighter skin, compared to 0.600 for darker skin. The PR value is 0.070 for lighter skin and -0.221 for darker skin. The large negative PR for darker skin (Z = 4.386, p $\leq 0.001$) reflects the model's reduced ability to balance precision and recall effectively in this subgroup, further emphasizing the AUC-PR results.
\end{itemize}

 \noindent The analysis reveals significant performance gaps between lighter and darker skin phototypes across key metrics, particularly precision, AUC-PR, and F1-score. These disparities highlight potential systemic biases in the algorithms trained with HAM10000 dataset, which lacks adequate representation of darker skin tones and fails to capture unique lesion characteristics prevalent in these populations. The PR values provide quantitative evidence of these gaps, consistently demonstrating that models perform substantially better for lighter skin tones. Although sensitivity and specificity metrics are less affected, precision and metrics balancing precision and recall (AUC-PR and F1-score) show significant negative PR values for darker skin phototypes. 

\begin{scriptsize}
\renewcommand{\arraystretch}{1} % Adjust row spacing for readability
\tabcolsep=0.09cm
\begin{table}
\centering
\footnotesize
\caption{Performance metrics for various models evaluated on Light and Dark subsets. Significance levels are indicated: * $p \leq 0.05$, ** $p \leq 0.01$, *** $p \leq 0.001.$}
\label{tab:metrics_performance}
\begin{tabular}{|l|c|c|c|c|c|c|c|cc|}
\hline
 &  & \multicolumn{8}{|c|}{BOSQUE test set} \\ \cline{3-10}
Model & HAM10000 & Overall & Light & Dark & \(\text{PR}_{\mathcal{M}_i}\) & \(\text{PR}_{\mathcal{M}_i}\) & Z-Statistic & \multicolumn{2}{|c|}{p-Value} \\ 
 &  &  (n=165) &  (n=107) & (n=58) & (Light)  & (Dark) &  & \multicolumn{2}{|c|}{} \\ 
\hline
\multicolumn{10}{|c|}{\textbf{Precision (Malignant)}} \\ \hline &  &  &  &  &  &  &  &  &  \\ 
ResNet50 & 0.751 & 0.780 & 0.897 & 0.484 & 0.118 & -0.296 & 5.874 & 4.26e-09 & *** \\
DenseNet121 & 0.734 & 0.692 & 0.815 & 0.415 & 0.123 & -0.277 & 5.243 & 1.58e-07 & *** \\
MobileNetV2 & 0.722 & 0.685 & 0.835 & 0.400 & 0.151 & -0.285 & 5.734 & 9.82e-09 & *** \\
EfficientNetV2B0 & 0.678 & 0.648 & 0.798 & 0.341 & 0.150 & -0.307 & 5.819 & 5.93e-09 & *** \\
VGG16 & 0.687 & 0.597 & 0.755 & 0.316 & 0.157 & -0.282 & 5.498 & 3.83e-08 & *** \\ \hline
\multicolumn{10}{|c|}{\textbf{Sensitivity}} \\ \hline &  &  &  &  &  &  &  &  &  \\ 
ResNet50 & 0.994 & 0.867 & 0.886 & 0.789 & 0.019 & -0.078 & 1.669 & 0.095 &  \\
DenseNet121 & 0.981 & 0.939 & 0.949 & 0.895 & 0.011 & -0.044 & 1.315 & 0.189 &  \\
MobileNetV2 & 0.868 & 0.908 & 0.899 & 0.947 & -0.009 & 0.039 & -1.074 & 0.283 &  \\
EfficientNetV2B0 & 0.919 & 0.827 & 0.848 & 0.737 & 0.022 & -0.090 & 1.736 & 0.083 &  \\
VGG16 & 0.968 & 0.969 & 0.975 & 0.947 & 0.005 & -0.022 & 0.913 & 0.361 &  \\ \hline
\multicolumn{10}{|c|}{\textbf{AUC-PR}} \\ \hline &  &  &  &  &  &  &  &  &  \\
ResNet50 & 0.929 & 0.901 & 0.949 & 0.749 & 0.048 & -0.152 & 3.762 & 1.68e-04 & *** \\
DenseNet121 & 0.917 & 0.871 & 0.916 & 0.740 & 0.044 & -0.132 & 3.053 & 0.002 & ** \\
MobileNetV2 & 0.801 & 0.782 & 0.878 & 0.528 & 0.096 & -0.254 & 4.994 & 5.93e-07 & *** \\
EfficientNetV2B0 & 0.804 & 0.692 & 0.843 & 0.372 & 0.152 & -0.320 & 6.192 & 5.93e-10 & *** \\
VGG16 & 0.794 & 0.715 & 0.865 & 0.496 & 0.151 & -0.219 & 5.136 & 2.81e-07 & *** \\ \hline
\multicolumn{10}{|c|}{\textbf{Specificity}} \\ \hline &  &  &  &  &  &  &  &  &  \\
ResNet50 & 0.670 & 0.642 & 0.714 & 0.590 & 0.072 & -0.052 & 1.625 & 0.104 &  \\
DenseNet121 & 0.646 & 0.388 & 0.393 & 0.385 & 0.005 & -0.003 & 0.104 & 0.917 &  \\
MobileNetV2 & 0.667 & 0.388 & 0.500 & 0.308 & 0.112 & -0.080 & 2.381 & 0.017 & * \\
EfficientNetV2B0 & 0.564 & 0.343 & 0.393 & 0.308 & 0.050 & -0.036 & 1.086 & 0.277 &  \\
VGG16 & 0.559 & 0.045 & 0.107 & 0.000 & 0.062 & -0.045 & 2.584 & 0.010 & ** \\ \hline
\multicolumn{10}{|c|}{\textbf{Accuracy}} \\ \hline &  &  &  &  &  &  &  &  &  \\
ResNet50 & 0.832 & 0.776 & 0.841 & 0.655 & 0.065 & -0.121 & 2.734 & 0.006 & ** \\
DenseNet121 & 0.813 & 0.715 & 0.804 & 0.552 & 0.089 & -0.163 & 3.424 & 6.16e-04 & *** \\
MobileNetV2 & 0.768 & 0.697 & 0.794 & 0.517 & 0.097 & -0.180 & 3.699 & 2.17e-04 & *** \\
EfficientNetV2B0 & 0.741 & 0.630 & 0.729 & 0.448 & 0.099 & -0.182 & 3.566 & 3.62e-04 & *** \\
VGG16 & 0.763 & 0.594 & 0.748 & 0.310 & 0.154 & -0.284 & 5.461 & 4.73e-08 & *** \\ \hline
\multicolumn{10}{|c|}{\textbf{AUC-ROC}} \\ \hline &  &  &  &  &  &  &  &  &  \\
ResNet50 & 0.945 & 0.847 & 0.877 & 0.776 & 0.030 & -0.071 & 1.697 & 0.090 &  \\
DenseNet121 & 0.930 & 0.833 & 0.823 & 0.810 & -0.010 & -0.023 & 0.208 & 0.835 &  \\
MobileNetV2 & 0.842 & 0.740 & 0.773 & 0.702 & 0.033 & -0.038 & 1.000 & 0.317 &  \\
EfficientNetV2B0 & 0.835 & 0.624 & 0.682 & 0.543 & 0.058 & -0.082 & 1.777 & 0.076 &  \\
VGG16 & 0.841 & 0.620 & 0.723 & 0.613 & 0.103 & -0.007 & 1.454 & 0.146 &  \\ \hline
\multicolumn{10}{|c|}{\textbf{F1-Score (Malignant)}} \\ \hline &  &  &  &  &  &  &  &  &  \\
ResNet50 & 0.855 & 0.821 & 0.892 & 0.600 & 0.070 & -0.221 & 4.386 & 1.15e-05 & *** \\
DenseNet121 & 0.840 & 0.797 & 0.877 & 0.567 & 0.081 & -0.230 & 4.512 & 6.42e-06 & *** \\
MobileNetV2 & 0.789 & 0.781 & 0.866 & 0.562 & 0.085 & -0.218 & 4.351 & 1.35e-05 & *** \\
EfficientNetV2B0 & 0.780 & 0.726 & 0.822 & 0.467 & 0.096 & -0.260 & 4.744 & 2.10e-06 & *** \\
VGG16 & 0.803 & 0.739 & 0.851 & 0.474 & 0.112 & -0.266 & 5.142 & 2.72e-07 & *** \\ \hline
\end{tabular}
\end{table}
\end{scriptsize}

\newpage
\section*{Discussion}

As AI continues to be involved in critical aspects of society, ensuring that predictive models are truly generalizable is no longer a purely technical concern; it is an ethical and practical imperative. Our work introduces \emph{Predictive Representativity} as a framework for assessing whether ML models generalize fairly across different subpopulations. Rather than relying only on traditional notions of representativity rooted in data sampling, PR shifts the focus to the alignment between true or ideal — and predicted distributions, allowing for a well-informed, outcome-driven approach to fairness evaluation.\\

\noindent The alignment between theoretical and fair distributions often remains purely conceptual. Moreover, data are often biased, preventing the accurate computation or estimation of these functions; however, this does not mean that they are nonexistent. In light of these considerations, this work also proposes a practical approach based on performance metrics, which are domain-specific and defined by experts to validate the representation of subpopulations. In the context of predictive algorithms, representativity cannot be claimed if these metrics do not achieve competitive performance levels comparable to a benchmark or the target population. \\

\noindent Through our case study on AI-driven skin cancer detection, we reveal systemic bias in the widely used HAM10000 dataset and some well-known deep learning models trained on it, which consistently favor lighter skin tones. Our analysis demonstrates that HAM10000 and some benckmark training strategies fail to capture the full spectrum of dermatological variations in diverse populations, leading to significant performance disparities. Although the dataset includes a proportional representation of darker skin cases relative to the general population, the detection of malignant lesions in these groups remains inadequate. This indicates a need for oversampling real examples on darker skin phototypes, as equal probability sampling does not ensure generalization. Our findings challenge the traditional notion of representativity as a simple scale-down of the target population and instead advocate for a predictivity-driven definition of representativity— one that prioritizes equitable model performance over static dataset composition.\\

\noindent Furthermore, while the prevalence of skin cancer is lower in individuals with darker skin tones compared to those with lighter skin, this should not justify their exclusion from dermatological AI models training. Ensuring representativity is critical, irrespective of disease prevalence, as underrepresentation in training datasets can lead to disproportionately poor diagnostic performance for these groups. A lower prevalence does not diminish clinical importance; in fact, as mentioned, when skin cancer occurs in darker-skinned individuals, it is often diagnosed at more advanced stages, resulting in worse prognoses and higher mortality rates. Therefore, it is essential to ensure representativity, regardless of what we consider to be prevalent or not.\\

\noindent In practice, ensuring fairness under distribution shifts requires more than proportional sampling; it calls for deliberate strategies like targeted oversampling and domain-specific data augmentation to enrich underrepresented patterns. Notably, disease prevalence differences should not be used to rationalize excluding minority groups from model development, especially when those groups face disproportionately severe consequences from errors. Instead, mitigating predictive inequities may demand weighting rare but critical subpopulations more heavily during training or model selection to safeguard their interests.\\

\noindent On the other hand, we acknowledge that the benchmark models considered in our case study do not achieve state-of-the-art performance; however, we caution that their continued and uncritical use risks obscuring deeply embedded racial biases within the training data. Benchmarks, often relied upon as neutral reference points for model comparison, may inadvertently reinforce unfair standards if the datasets on which they are built are inherently biased. Although data sets like HAM10000 are widely used in academic research, their unchecked influence on AI-driven medical technologies may reinforce existing social injustices rather than mitigate them. Our findings confirm the racial bias in HAM10000 previously identified by \cite{morales2024insight} and emphasize the need to reassess any claims of representativeness in its documentation. More broadly, our analysis highlights the pressing need to shift AI fairness efforts away from a narrow focus on dataset composition and toward a more rigorous evaluation of predictive equity and conditional distribution alignment across diverse populations. Achieving this requires more inclusive and context-aware data collection, responsible and transparent documentation, targeted algorithmic fairness interventions, and continuous performance assessment beyond general accuracy metrics. \\

\noindent  From a mere technical point of view, our PR framework can also be understood as a measure of robustness when applied to specific sub-populations and tailored to a defined prediction task. However, it must be used with caution. The primary role of PR is for external validation, not model training. We emphasize that PR is most effective as a post-hoc auditing metric and diagnostic tool, rather than as a direct optimization objective during training. Optimizing a model solely to minimize PR (i.e., to equalizeperformance across subgroups at all costs) could induce overfitting to the idiosyncrasies of particular subgroups and undermine generalization. In other words, aggressively
forcing a model to “chase” parity on PR might trade off overall calibration and stability for one context at the expense of performance in others. For example, leveraging the BOSQUE Test set to equalize performance across skin tones may expand the scope of training data but does not address the fundamental challenge of selecting cases that are truly representative for predictive accuracy —enriching HAM10000 training data with the BOSQUE Test set does not ensure reliable model deployment in a Colombian context—. The goal should be on identifying examples that enhance model reliability in real-world applications, rather than merely achieving training balance or addressing demographic shifting. Ensuring strong transferability —the ability of a model trained on one dataset to perform well in different but related settings— in AI models is crucial in this regard. A model that generalizes effectively across different skin types and geographic regions will be more reliable in real-world clinical settings. Instead of narrowly optimizing for one dataset, fairness evaluations should include assessments of a model’s adaptability to varied contexts and its resilience to shifts in population distributions.\\

 \noindent Predictive representativeness also expands the discussion beyond the traditional notion of “representative samples” by emphasizing that representativity cannot be assessed in isolation—it must always be contextualized. A sample's representativity is meaningful only when its intended purpose and the specific subpopulation it aims to represent are clearly defined. In other words, are representative they \emph{for whom} and \emph{for what purpose}? Without this clarity, representativity becomes an ambiguous concept that risks being misinterpreted or misapplied.  In our case study, and more broadly in ML,  a dataset that appears representative in terms of overall demographic distribution may still fail to provide meaningful predictive insights if it does not adequately capture the structural patterns necessary for accurate decision-making across subpopulations. Assessing representativity without considering the specific predictive task can lead to erroneous assumptions about model performance. For instance, a dataset that is well-representative for diagnosing one medical condition may not be representative for another, even if drawn from the same population. This underscores the importance of defining representativity in alignment with both the prediction objective and the characteristics of the population at risk.\\

\noindent Our proposed PR framework has implications that extend well beyond the dermatology case study. As a concrete and interpretable metric for post-hoc fairness auditing, PR can be applied to any domain where one needs to quantify outcome disparities across demographic or context-defined subgroups. In fields ranging from medical diagnostics to finance and public policy, it offers a check on whether improvements in aggregate accuracy might be coming at the cost of worse outcomes for marginalized subpopulations. \\

\noindent Crucially, our results highlight the inevitable trade-offs between global performance optimization and subgroup equity that organizations must navigate. Recent research continues to underscore this fairness–accuracy tension, as achieving low disparity across groups often entails some sacrifice in overall accuracy or efficiency \citep{rabonato2024systematic}. Rather than viewing this as an undesirable byproduct to ignore, stakeholders should treat it transparently as a design consideration—what degree of performance trade-off is acceptable to ensure that no group is left behind? Addressing these compromises in an open, principled manner will be critical for cultivating trust in AI systems. For example, models that maximize average utility at the expense of minority groups risk entrenching existing inequities, whereas those developed with fairness constraints may slightly reduce headline performance metrics but ultimately provide more reliable and just outcomes across diverse environments. By surfacing such considerations, PR encourages a more nuanced conversation about model objectives, one that goes beyond one-size-fits-all metrics and towards inclusive, context-sensitive criteria for equitable decision-making.\\

\noindent Furthermore, our study contributes a formal \textit{External Transportability Criterion} grounded in the PR metric, which connects fairness and robustness under shifting contexts. We posit that for a model to be considered \textit{fairly transportable} from a source population $\mathcal{P}$ to a new target population $\mathcal{P}'$, the absolute value of its PR should remain within an acceptable tolerance $\varepsilon$ on the target domain; that is, $|PR(\mathcal{P}'; S', A)| \leq \varepsilon$. This criterion operationalizes the notion of generalizing fairness across deployment contexts by setting a concrete threshold on allowable performance disparity between populations.\\

\noindent Although this approach aligns with recent efforts to ensure robustness under distributional shift, it also reveals key limitations of conventional validation practices. In our case study, even benchmark skin lesion classifiers that performed well on the original test set failed to meet a standard PR tolerance when evaluated on a demographically distinct dataset. This outcome underscores the limitations of relying on black-box models for fairness assessments: Their opaque decision boundaries can mask subgroup-specific failures, making it difficult to trace and correct inequities when models are deployed in new contexts. As a result, post-hoc audits may be insufficient unless interpretability is integrated into the model development lifecycle.\\

\noindent To enhance transportability assessment, future research should draw more explicitly on causal inference frameworks, particularly those based on transportability theory \citep{pearl2011transportability}. These approaches offer formal tools to determine when fairness properties can be expected to generalize, based not only on statistical correlations, but on structural assumptions about the underlying data-generating processes. Embedding such causal reasoning into fairness diagnostics could enable more principled evaluations of whether a model's equitable behavior in one setting is likely to hold in another, thereby bridging the gap between empirical performance and real-world reliability.\\

\noindent  It is important to note that, ideally, fairness and representativity audits should be conducted \emph{before} deploying a model, using whatever data are available from the source domain. In our case, such proactive auditing at the source was limited by the absence of explicit skin tone or ethnicity labels in the HAM10000 training dataset. This limitation reflects a broader challenge in fairness auditing: many real-world datasets lack annotations for key sensitive attributes, making it difficult to quantify subgroup performance disparities prior to deployment \citep{mittal2024responsible}. Despite this hurdle, our evaluation on the external BOSQUE Test set effectively served as a post-hoc audit, revealing that ostensibly strong overall accuracy of the classifiers concealed serious subgroup-specific failures. The stark drop in precision for darker-skinned patients,unseen during source validation, illustrates how an AI model can appear well behaved on aggregate metrics, yet still propagate harm when applied in a different demographic context. This underscores the danger of relying solely on overall accuracy or naive data set “representativeness” checks without deeper interrogation of model behavior.\\

\noindent On the other hand, we chose not to rely on simulation-based evaluations. Our decision was grounded in both methodological and ethical considerations. First, the aim of PR is to expose how fairness — or lack thereof — emerges under real-world demographic shifts. Synthetic simulations, while useful in controlled benchmarking tasks, often require parametric assumptions or stylized subgroup definitions that risk obscuring the structural and institutional biases embedded in real clinical data. Using two datasets, HAM10000 and BOSQUE Test set, we evaluated fairness failures as they manifest in actual deployment contexts, lending ecological validity to our findings. Moreover, the analytical formulation of PR and the External Transportability Criterion does not depend on stochastic approximations; instead, it enables deterministic and interpretable assessments of subgroup-level predictive equity. In fairness auditing, particularly in health-related AI, empirical evidence drawn from real-world populations carries greater diagnostic and ethical weight than simulated outcomes, which may understate or mischaracterize lived disparities.\\

\noindent In the future, we advocate that predictive representativity analysis (and analogous subgroup fairness evaluations) become a standard component of model validation and risk assessment. This integration would complement traditional performance metrics with explicit checks for subgroup equity, thereby strengthening the robustness of the model under distributional shifts and providing greater assurance of fairness before real-world deployment. In summary, by illuminating hidden performance gaps and prompting targeted mitigation, the PR framework helps ensure that “success” in machine learning is defined not just by overall accuracy, but by the model’s ability to perform equitably across the spectrum of populations it claims to serve.

\section*{Conclusion}

Achieving equitable performance in AI systems demands more than balanced training data—it requires outcome-aware, context-sensitive validation. Through the introduction of \emph{Predictive Representativity}, we provide a rigorous framework to quantify disparities in predictive behavior across subpopulations, revealing fairness failures that conventional aggregate metrics may obscure.\\

\noindent Our findings challenge the assumption that proportional sampling ensures algorithmic fairness. Even when darker skin phototypes were proportionally present in the training set —proportional to prevalence—, classifiers exhibited significant performance gaps under subgroup distribution shifts, reflecting poor subgroup shift robustness. These disparities were only revealed through granular, post-hoc evaluation, echoing prior reports that model-centric validation alone often misses critical fairness issues. In particular, we empirically verified racial bias in the widely used HAM10000 dataset—despite its apparent diversity, it fails to support equitable generalization for darker skin tones.\\

\noindent PR offers a structured pathway toward such granular audits—operationalizing fairness as a measurable alignment between predictions and ground truth— or ideal— across demographic groups. Combined with the \textit{External Transportability Criterion}, this framework enables stakeholders to assess whether fairness generalizes beyond the source population, a prerequisite for trustworthy deployment in real-world, demographically diverse settings.\\

\noindent By reframing representativity as an outcome-level property, PR invites a shift from passive dataset documentation toward active, iterative fairness evaluation. As AI systems increasingly impact public health and societal decisions, such tools will be indispensable for ensuring that predictive success includes—and serves—all populations equitably.\\

%\section*{conclusion}

%Our findings echo prior reports that exclusively model-centric validation can miss critical fairness issues which only become evident under more granular, context-specific evaluation. \\

%Our findings challenge the misconception that proportional sampling of demographic groups automatically guarantees algorithmic fairness. Even with a balanced representation of darker skin phototypes in the training data, we observed significant performance disparities on the shifted test distribution. This scenario exemplifies a subgroup distribution shift (a form of covariate shift focused on sensitive subpopulations) under which the model exhibited poor subgroup shift robustness: accuracy and precision degraded notably for the minority subgroup. Such results underscore that a dataset can be statistically “balanced” yet still fail to ensure fairness under covariate shift, due to a lack of subgroup robustness in outcomes.\\

\subsection*{Acknowledgements}
The authors gratefully acknowledge the support of Universidad El Bosque for facilitating the collection and curation of the BOSQUE Test set. We extend our special thanks to dermatologists Dr. Alejandra Jaramillo Arboleda and Dr. Maria Juliana Sánchez-Zapata for their crucial roles in the clinical coordination and image acquisition process.\\

\noindent We are also deeply thankful to Professor Jorge Humberto Mayorga, Francisco Perez, and Dr. Anne-Déborah Bouhnik for their insightful feedback and critical discussions, which significantly improved the framing and clarity of the manuscript. AI was used for writing assistance. 

\subsection*{Author Contributions}
\emph{Conceptualization:} Andrés Morales-Forero, Ronald Herrera, Lili J. Rueda, Samuel Bassetto, and Eric Coatanea.
\emph{Methodology and study design:} Andrés Morales-Forero and Lili J. Rueda.
\emph{Coding and computational analysis:} Andrés Morales-Forero.
\emph{Validation and formal analysis of results:} Andrés Morales-Forero and Lili J. Rueda.
\emph{Writing – original draft preparation:} Andrés Morales-Forero.
\emph{Writing – review and editing: } Andrés Morales-Forero, Lili J. Rueda, Ronald Herrera, Samuel Bassetto, and Eric Coatanea.
\emph{Supervision:} Samuel Bassetto and Eric Coatanea.\\

\noindent This project was a multidisciplinary effort that combined clinical dermatology, epidemiology, and machine learning expertise. All authors have read and approved the final version of the manuscript.

\subsection*{Conflict of interest and funding} \noindent The authors declare no competing interests. No external funding influenced the study design, data analysis, or interpretation.

\subsection*{Data protection and privacy} 

\noindent All analyses were performed on secured, access-controlled servers. The research team received only pseudonymised data; no attempt was made to re-identify participants, and results are reported exclusively in aggregated form.

\subsection*{Data Availability Statement} 

\noindent  The BOSQUE Test Set, an anonymized and depurated subset of the original BOSQUE dataset, is publicly available through the Harvard Dataverse repository: \href{ https://doi.org/10.7910/DVN/AQEPIN}{https://doi.org/10.7910/DVN/AQEPIN}. \\

\noindent The original data set used in the study, which includes comprehensive prospectively collected clinical images and detailed metadata, remains restricted due to institutional policies and Colombian health data privacy regulations. Access to the full version may be granted upon reasonable request and with prior approval from Universidad El Bosque and the corresponding ethics committee. \\

\noindent  The HAM10000 data set is also publicly available and can be accessed via the Harvard Dataverse repository: \href{ https://doi.org/10.7910/DVN/DBW86T}{https://doi.org/10.7910/DVN/DBW86T}. \\

\noindent All source code and evaluation scripts associated with this study are available in the GitHub repository: \href{https://github.com/jamorafo/DermAlgoFairness}{https://github.com/jamorafo/DermAlgoFairness}.

\subsection*{Ethics Statement}

This research complies with all relevant national regulations and with the ethical principles of the Declaration of Helsinki. Two data sources were analysed:

\begin{enumerate}
    \item \textbf{HAM10000 public reference set}\\
    The HAM10000 dermoscopic image repository is openly available and fully de-identified \citep{tschandl2018ham10000}. Because all images are anonymised at source, no additional institutional-review-board (IRB) approval was required for their secondary use in this study.
    
    \item \textbf{BOSQUE Test set}\\
    All BOSQUE images and metadata were prospectively collected and pseudonymised \emph{in-house} by clinical staff and researchers at \textbf{Universidad El Bosque, Bogotá, Colombia} before any transfer or analysis.  
    The study protocol — \textit{“Características operativas de una prueba diagnóstica basada en un algoritmo de inteligencia artificial entrenado con las imágenes del HAM10000 para el diagnóstico de lesiones pigmentadas en fototipos claros y oscuros”} — received full approval from the Institutional Review Board (Comité de Ética en Investigación, CEI) of the Subred Integrada de Servicios de Salud Norte E.S.E.\ (Acta 81, 06 December 2023; project code SNCEI-205).  
    Written informed consent was obtained from all participants, and no directly identifying information was retained.
\end{enumerate}

\newpage
\begin{appendices}

\section*{Comparison of Predictive Representativity with Related Concepts}
\label{appendix:PR_comparisson}

\noindent To clarify the unique role of Predictive Representativity within the broader ecosystem of ML evaluation frameworks, the Table \ref{tab:pr_comparison} provides a comparative summary of PR and commonly related concepts. This comparison helps delineate the theoretical and practical boundaries between fairness, robustness, transportability, and representativity.

\begin{table}[h!]
\caption{Conceptual comparison between Predictive Representativity and other model evaluation paradigms.}
\label{tab:pr_comparison}
\centering
\scriptsize
\begin{tabular}{|p{3.5cm}|p{5cm}|p{5cm}|}
\hline
\textbf{Concept} & \textbf{Core Focus} & \textbf{Difference from PR} \\
\hline
Model Portability & Technical compatibility with different environments & Focuses on system-level deployability, not predictive equity or fairness diagnostics. \\
\hline
Model Adaptability & Capacity to retrain or fine-tune under new conditions & Involves learning dynamics; PR assumes static models and audits fairness at prediction time. \\
\hline
Internal Transportability & Validity within the original training population & PR extends this by assessing subgroup fairness, not just aggregate performance. \\
\hline
External Transportability & Generalization of model fairness to new populations & PR formalizes this via a quantitative fairness threshold ($\epsilon$) across subgroups and contexts. \\
\hline
Model Robustness & Stability under data perturbations or shifts & PR captures a specific form: group-conditional prediction stability. \\
\hline
Demographic Parity & Equal prediction rates across groups & Unlike PR, does not reference ground truth; PR can tolerate predictive differences if label distributions differ. \\
\hline
Equalized Odds & Equal true/false positive rates across groups & PR provides a divergence-based fairness lens without assuming label parity. \\
\hline
Calibration by Group & Consistency between predicted probabilities and observed outcomes & PR generalizes beyond calibration by incorporating a framework for inference evaluation. \\
\hline
Coverage (Data Representativity) & Inclusion of all relevant groups in dataset & PR goes further: group inclusion alone doesn’t ensure fair predictions; evaluates performance after inclusion. \\
\hline
Generalization & Performance on unseen data & PR detects when generalization fails for specific subgroups, especially under demographic shifts. \\
\hline
Transfer Learning & Applying models to related domains/tasks & Transfer may aid PR but doesn't guarantee subgroup equity; PR audits fairness post-transfer. \\
\hline
Bias Auditing & Post-hoc fairness evaluation & PR is a rigorous metric that operationalizes bias detection using predictive divergence and fairness gaps. \\
\hline
\end{tabular}
\end{table}

\end{appendices}

\bibliographystyle{apacite}

\bibliography{references}

\end{document}